\def\year{2022}\relax
\documentclass[letterpaper]{article} 
\usepackage{aaai22}  
\usepackage{times}  
\usepackage{helvet}  
\usepackage{courier}  
\usepackage[hyphens]{url}  
\usepackage{graphicx} 
\usepackage{natbib}  
\usepackage{caption} 
\DeclareCaptionStyle{ruled}{labelfont=normalfont,labelsep=colon,strut=off} 
\usepackage{array}
\usepackage{bbm}
\usepackage[ruled,vlined]{algorithm2e}
\usepackage{multirow}
\usepackage{amsmath}
\usepackage[normalem]{ulem}
\useunder{\uline}{\ul}{}

\usepackage{newfloat}
\nocopyright
\usepackage[breaklinks=true,colorlinks,bookmarks=false]{hyperref}

\title{Self-supervised Representation Learning Framework for Remote Physiological Measurement Using Spatiotemporal Augmentation Loss}
\author{
    Hao Wang, 
    Euijoon Ahn, 
    Jinman Kim
}
\affiliations{
    The University of Sydney, Sydney, NSW, Australia\\
    hwan7885@uni.sydney.edu.au, \{euijoon.ahn, jinman.kim\}@sydney.edu.au
    
}

\begin{document}

\maketitle

\begin{abstract}
    Recent advances in supervised deep learning methods are enabling remote measurements of photoplethysmography-based physiological signals using facial videos. The performance of these supervised methods, however, are dependent on the availability of large labelled data. Contrastive learning as a self-supervised method has recently achieved state-of-the-art performances in learning representative data features by maximising mutual information between different augmented views. However, existing data augmentation techniques for contrastive learning are not designed to learn physiological signals from videos and often fail when there are complicated noise and subtle and periodic colour/shape variations between video frames. To address these problems, we present a novel self-supervised spatiotemporal learning framework for remote physiological signal representation learning, where there is a lack of labelled training data. Firstly, we propose a landmark-based spatial augmentation that splits the face into several informative parts based on the Shafer’s dichromatic reﬂection model to characterise subtle skin colour fluctuations. We also formulate a sparsity-based temporal augmentation exploiting Nyquist–Shannon sampling theorem to effectively capture periodic temporal changes by modelling physiological signal features. Furthermore, we introduce a constrained spatiotemporal loss which generates pseudo-labels for augmented video clips. It is used to regulate the training process and handle complicated noise. We evaluated our framework on 3 public datasets and demonstrated superior performances than other self-supervised methods and achieved competitive accuracy compared to the state-of-the-art supervised methods. Code is available at \url{https://github.com/Dylan-H-Wang/SLF-RPM}.
\end{abstract}

\section{Introduction}

Physiological signals are critical indicators for human cardiovascular activities such as heart rate (HR), respiration frequency (RF), heart rate variability (HRV) and blood pressure (BP) \cite{XU202155}. These signals are commonly used to monitor the wellness of patients \cite{Avram2019RealworldHR}. Traditionally, the Electrocardiography (ECG) and Photoplethysmography (PPG) are used to measure physiological signals, and both of them rely on the availability of cuff-based equipment which requires direct contact to human skin. This constrains the application of monitoring and estimation process in an unobtrusive and concomitant way with ubiquitous devices (e.g., smartphone cameras, webcams) \cite{McDuff}. In recent years, non-contact video-based remote physiological measurement (RPM) has been of great interest. Remote photoplethysmography (rPPG), using facial videos, has been introduced to overcome the limitation of conventional contact-based measurement approaches. In rPPG, signals are measured based on colour fluctuations on human skin, which are caused by the variations of blood volume during cardiac cycle \cite{de2013robust}. 

Several recent studies using supervised deep learning methods \cite{Yu3, Qiu, Niu4, Lee, Li2018TheOD} have shown promising results to remotely estimate physiological signals. They, however, remain problematic because of their dependency on the availability of large-scale labelled training data. The annotation of large-scale data is costly, slow and requires medical equipment. Researchers have employed various approaches to help solve these challenges including transferring knowledge across different domains and fine-tuning those knowledge with a relatively smaller amount of labelled image data. For example, Niu et al. \shortcite{Niu2} used a model that was pre-trained using large labelled natural images (i.e., ImageNet) and fine-tuned this knowledge using rPPG facial videos. Another approach is to create synthetic physiological signals to increase the number of training videos \cite{Niu1,Condrea2020InSO}. However, such approach is limited by the domain shift between original and synthetic data. An alternative approach is to use self-supervised learning (SSL) to learn and extract image features from unlabelled data. Many recent SSL methods commonly applied the concept of contrastive learning and have achieved state-of-the-art (SOTA) performances in unsupervised image/video representation learning \cite{Ahn2020UnsupervisedDA, Oord2018RepresentationLW, Han2019VideoRL, Qian2020SpatiotemporalCV, Ahn2021ASG}. In these approaches, videos are transformed using standard data augmentation techniques such as frame cropping, resizing, colour jittering and frame re-ordering to produce different views. Invariant video features are then learnt by maximising mutual information between different views in a contrastive manner. These standard augmentation techniques, however, are mainly limited to learn features for action recognition tasks, where large variations among the human anatomy can be modelled, and they are not designed to capture the subtle colour fluctuations on human skin. Many of RPM studies also transformed the video frames (3D) into 2D spatiotemporal map for subsequent 2D convolutional neural networks (CNNs) training \cite{Niu2, Qiu, Lee}. However, this transformation potentially neglect nature information contained in original inputs \cite{Yu2}.

In this paper, we present a new Self-supervised Learning Framework for Remote Physiological Measurement (SLF-RPM). We propose a landmark-based spatial augmentation using Shafer’s dichromatic reﬂection model~\cite{Wang2017AlgorithmicPO} to effectively capture the colour fluctuations on human faces. We also propose a sparsity-based temporal augmentation that characterise periodic colour variations using Nyquist–Shannon sampling theorem~\cite{NyquistCertainTI} to exploit rPPG signal features. We further formulate a new loss function using the pseudo-labels derived from our augmentations. It regulates the training process of contrastive learning and handles complicated noise. We evaluated our framework by comparing with other SOTA supervised and SSL approaches using 3 public datasets and conducted ablation studies to demonstrate the effectiveness of our SLF-RPM framework.

\section{Related Work}

\subsection{Remote Physiological Measurement}

The application of analysing rPPG from camera-captured videos was first proposed by Verkruysse et al.~\shortcite{verkruysse2008remote}. In early work, many studies have manually designed hand-crafted signal features to characterise the rPPG signals. For example, Poh et al.~\shortcite{poh2010non,poh2010advancements} used independent component analysis (ICA) with RGB colour sequences to estimate HR signals. Similarly, some methods used chrominance features reflected from the human skin ~\cite{wang2014exploiting,wang2016algorithmic}. Although these hand-crafted features have shown promising performance, they are required to manually select region of interest (ROI), detect and process skin-pixels signals. This is challenging or even quixotic to be implemented in practical settings. In recent years, deep learning methods based on CNNs ~\cite{tulyakov2016self,hsu2017deep,Niu1} have been developed to overcome such limitations and they have been shown to effectively capture minor colour variations and extract rPPG signals. For example, Niu et al.~\shortcite{Niu2} proposed RhythmNet to transform each video clip into a 2D feature map and fed it into a CNN to estimate rPPG signals. Yu et al.~\shortcite{Yu3} proposed an end-to-end CNN model, i.e, rPPGNet to reconstruct rPPG signals from highly-compressed videos. Lee et al.~\shortcite{Lee} applied the concept of meta-learning using 2D CNN coupled with bidirectional long short-term memory to learn spatiotemporal features and enable faster inference adaption. Moreover, Lu et al.\shortcite{LuDualGANJB} proposed Dual-GAN to model rPPG and noise signals to improve model robustness. The performance of these methods, however, were dependent on the labelled training data. Few unsupervised work were introduced \cite{Condrea2020InSO, Bobbia2019UnsupervisedST} and showed limited performances.

\subsection{Self-Supervised Video Representations Learning}

In recent years, self-supervised video representations learning methods have achieved promising results on action recognition task \cite{Han2019VideoRL, Han2020MemoryaugmentedDP, Jing2018SelfsupervisedSF}. For instance, Misra et al.\shortcite{Misra2016ShuffleAL} learnt video representations by classifying shuffled video clips. Similarly, Jenni et al.\shortcite{Jenni2020VideoRL} generated a set of temporal transformations and constructed a 3D CNN to recognise them. The essential concept with these approaches is to use data augmentations to learn invariant features along spatial and temporal axes. These approaches primarily dealt with object interactions, optical flows, synchronised audios and object tracking. They, however, were not designed to capture the subtle facial colour changes for estimating rPPG signals. 

\subsection{Data Augmentation}

Spatial augmentations\footnote{We also consider appearance transformation (such as colour distortion, Gaussian blur) as spatial augmentation.} are commonly used in both supervised and SSL to cover a wider and diverse data distribution, and they have shown to be effective in extracting discriminative image features~\cite{Shorten2019ASO,Han2020SelfsupervisedCF,Chen2020ImprovedBW}. However, existing RPM studies~\cite{Yu1, Qiu, Spetlik, Niu3, Chen2018DeepPhysVP} did not apply spatial augmentations to frames directly, and the most close one is proposed by Niu et al.~\shortcite{Niu4} which horizontally and vertically ﬂip the processed \textit{spatiotemporal map} (a representation to compact the 3D video into 2D). 

Motion statistics are essential in representing video information and play a key factor on the success of action recognition~\cite{Wang2019SelfSupervisedSR}. These video dynamics are closely related with temporal information where each action class has a common pattern that can be observed in subsequent frames~\cite{Jenni2020VideoRL}. For instance, many studies \cite{Brattoli2017LSTMSF,Fernando2017SelfSupervisedVR,Lee2017UnsupervisedRL,Xu2019SelfSupervisedSL,Kim2019SelfSupervisedVR} proposed to use image frame re-ordering or optical flows to augment temporal data. These augmentation techniques were, however, not designed to estimate physiological signals in facial videos.  

\section{Method}

\begin{figure*}
\begin{center}
    \includegraphics[width=.75\linewidth]{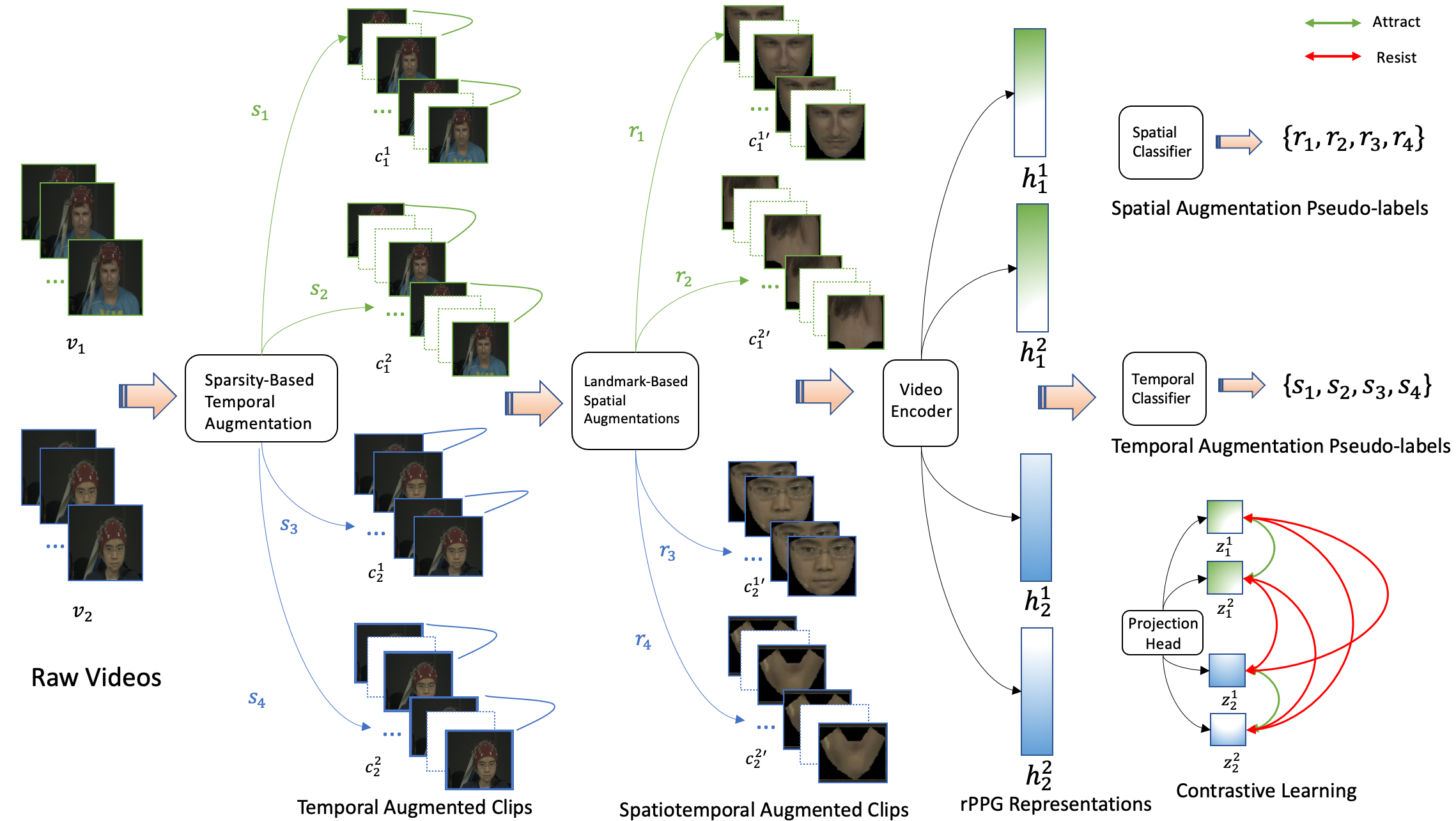}
\end{center}
   \caption{\textbf{Overview of our SLF-RPM.} Raw videos are transformed by the sparsity-based temporal augmentation and landmark-based spatial augmentation to generate different views which are then used in our contrastive learning. Simultaneously, pseudo-labels derived from our augmentations are used to constrain the learning process. Positive samples are denoted by the same subscripts and different superscripts e.g., $z_i^1, z_i^2$.}
\label{fig:framework}
\end{figure*}

\subsection{RPM Representation Learning Framework}

The overview of our SLF-RPM is illustrated in Fig.~\ref{fig:framework} and detailed algorithm in appendix~\ref{alg}. Suppose we have randomly sampled $N$ raw videos from the dataset. We first apply our sparsity-based temporal and landmark-based spatial augmentations to generate $2N$ samples in total. Two augmented clips $c_i^{1'}, c_i^{2'}$ originated from the same video $v_i$ are then fed into the \textit{video encoder} to obtain corresponding video features $h_i^1, h_i^2$. Both of them are then mapped by the \textit{projection head} into the space $z_i^1, z_i^2$ where contrastive loss \cite{Chen2020ASF} is applied to maximise mutual information. These two feature vectors are regarded as positive samples and rest of $2(N-1)$ samples within the same mini-batch are considered as negative samples. We then generated pseudo-labels based on applied augmentations of the video $v_i$ and used two classifiers (i.e., single fully-connected (FC) layer) trained on $h_i^1, h_i^2$ to identify the transformation $\{s_1, r_1\}$ and $\{s_2, r_2\}$ (see Sec.~\ref{pseudo-label}).

\subsection{Video Encoder and Projection Head}


\begin{table}
\begin{center}
\begin{tabular}{m{0.1\linewidth} |
>{\centering} m{0.35\linewidth} |
>{\centering\arraybackslash} m{0.35\linewidth}}
\hline
Layer Name & Layer Structure & Output Size \newline $(C \times D \times H \times W)$ \\
\hline\hline
inputs & N/A & $3 \times 30 \times 64^2$ \\
\hline
$\text{conv}_1$ & kernel size $7^3$ \\ temporal stride $1$ \\ spatial stride $2$ & $64 \times 30 \times 32^2$ \\
\hline
max pool & kernel size $3^3$ \\ stride $2$ & $64 \times 15 \times 16^2$ \\
\hline
$\text{conv}_2$ & basic block 2 & $64 \times 15 \times 16^2$ \\
\hline
$\text{conv}_3$ & basic block 2 & $128 \times 8 \times 8^2$ \\
\hline
$\text{conv}_4$ & basic block 2 & $256 \times 4 \times 4^2$ \\
\hline
$\text{conv}_5$ & basic block 2 & $512 \times 2 \times 2^2$ \\
\hline
\multicolumn{3}{c}{global average pool} \\
\hline
\end{tabular}
\end{center}
\caption{\textbf{Video encoder of SLF-RPM: 3D ResNet-18.} Each convolutional layer is followed by a batch normalisation layer and a ReLU layer. The spatiotemporal down-sampling is only applied to $\text{conv}_3$, $\text{conv}_4$ and $\text{conv}_5$ with stride of $2$. $C$, $D$, $H$ and $W$ denote
channels, frames, height and width.}
\label{tab:encoder}
\end{table}

We adopted 3D ResNet architecture \cite{hara3dcnns} as the video encoder which is shown in Table~\ref{tab:encoder}. The 3D architecture allows to learn spatial and temporal information at the same time. Each input was converted into a flatten feature vector $h$ and then fed into the projection head, which is a multi-layer perceptron (MLP) in our experiment, to obtain the final encoded feature vector (i.e., $z$ in Eq.~\ref{equ1}). The projection head is removed during the evaluation process, and the feature vector $h$ from the video encoder is used directly as RPM representations to make final predictions. 

\subsection{Preliminary Background: Skin Reflection Model}\label{skinModel}
 
According to Shafer’s dichromatic reﬂection model (DRM)~\cite{Wang2017AlgorithmicPO}, light source has a constant spectral composition with varying intensities and therefore the variation of skin reflections over the time are measured based on body motions (specular variations) and pulse-induced subtle colour changes (diffuse reflection) where only diffuse reflection contains rPPG-related information. Using DRM, we can then define the skin reflection model for the image sequence along the time by
\begin{equation}
    C_k(t) = I(t) \cdot (v_s(t) + v_d(t)) + v_n(t)
    \label{equ2}
\end{equation}
where $C_k(t)$ is the k-th skin pixel of RGB values; $I(t)$ denotes the light intensity level from the light source which is regulated by specular reflection $v_s(t)$ and diffuse reflection $v_d(t)$; $v_n(t)$ is the noise from camera sensor; $t$ is the time step. We can further decompose $v_s(t)$ and $v_d(t)$ by
\begin{equation}
    v_s(t) = u_s \cdot (s_0 + s(t))
    \label{equ3}
\end{equation}
where $u_s$ is the unit colour vector of the light spectrum, $s_0$ and $s(t)$ are the stationary and varying parts of specular reﬂections, i.e., $s(t)$ captures motions.
\begin{equation}
    v_d(t) = u_d \cdot d_0 + u_p \cdot p(t)
    \label{equ4}
\end{equation}
where $u_d$ denotes the unit colour vector of the skin pixel; $d_0$ refers to the stationary reﬂection strength; $u_p$ refers to the relative signal strengths in RGB channels; $p(t)$ refers to the rPPG signal. Given the defined notation above, we can rewrite the Eq.~\ref{equ2} using Eq.~\ref{equ3} and ~\ref{equ4} by
\begin{equation}
    C_k(t) = I(t) \cdot (u_s \cdot (s_0 + s(t)) + u_d \cdot d_0 + u_p \cdot p(t)) + v_n(t)
    \label{equ5}
\end{equation}
where our aim is to calculate $p(t)$ from $C_k(t)$.

\subsection{Landmark-Based Spatial Augmentation}\label{spatialAug}

\begin{figure}
\begin{center}
   \includegraphics[width=.7\linewidth]{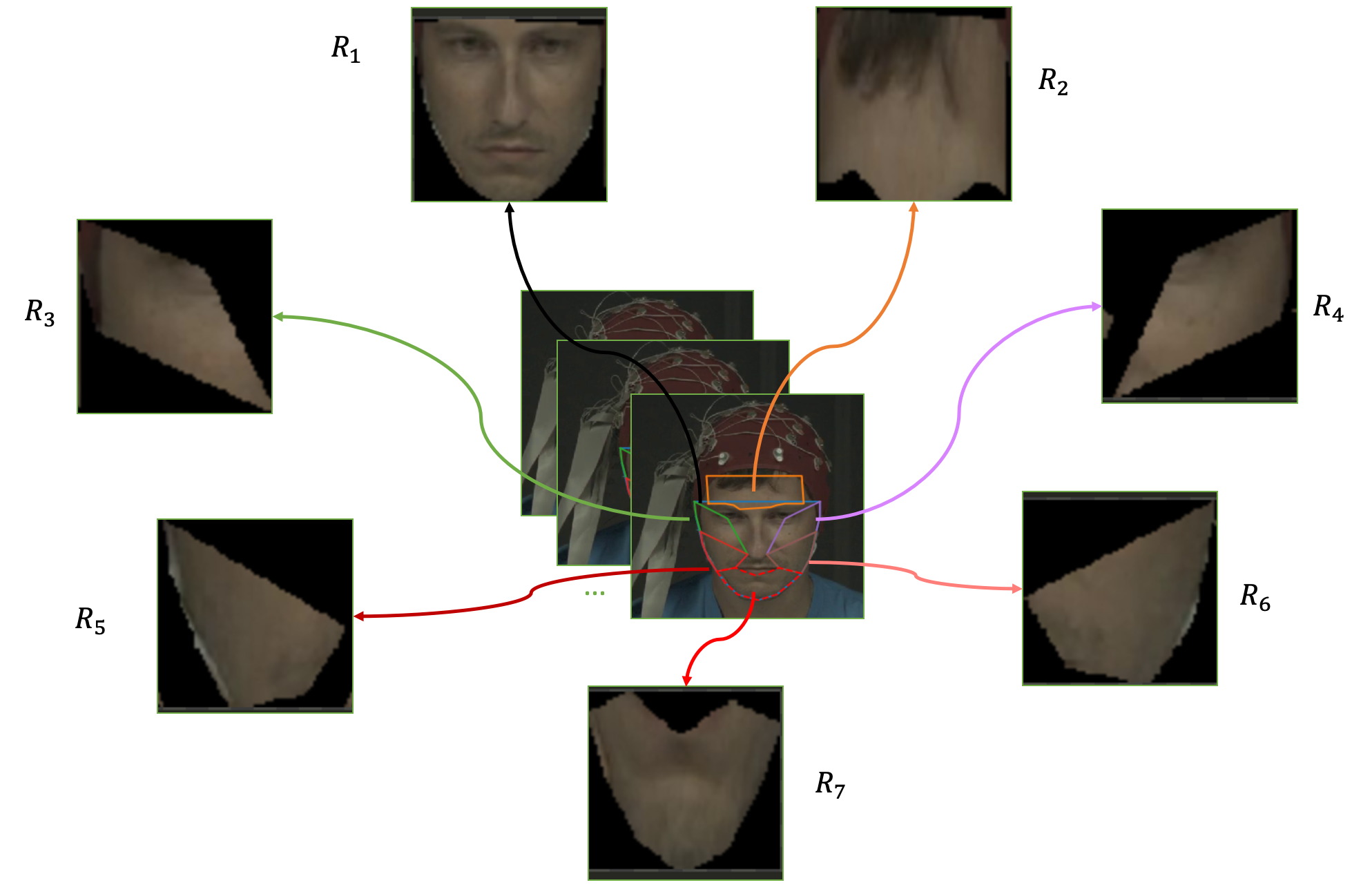}
\end{center}
   \caption{\textbf{Illustration of landmark-based spatial augmentation.} Based on detected facial landmarks, we define 7 ROI areas $\{R_1, R_2, R_3, R_4, R_5, R_6, R_7\}$.}
\label{fig:spatialAug}
\end{figure}

We defined two criteria for our landmark-based spatial augmentation: 1) the relationship among colour channels of each pixel needs to be considered so that any pixel value changes (e.g., colour jitter) are not allowed; and 2) the underlying colour variations among different views are similar. From Eq.~\ref{equ5}, the rPPG information contained in videos can be maximised by forcing $u_d$ to dominate the equation, and hence the diffuse reflection $v_d(t)$ can be approximated by skin pixel value $C_k(t)$. As such, we selected and cropped each frame into several facial parts according to face landmark locations~\cite{Bulat2017HowFA} as shown in Fig~\ref{fig:spatialAug}. The selection of ROIs considers two factors: 1) the movements of eyes and mouth are more rapid than other parts of face which put more weights on the specular reflection $v_s(t)$ \cite{Li2014RemoteHR}; and 2) facial parts in the same video with similar skin colour $C_k(t)$ should contain similar signals $p(t)$. Theoretically, removal of non-facial areas filters noise from the background which reduces the weight of $v_n(t)$ in Eq.~\ref{equ5}. Informative selected ROIs ensure the dominance of diffusion reflections $v_d(t)$ and different ROI sequences from the same facial video contain similar rPPG signals which are used as positive samples (i.e., ($z_i, z_j$) in Eq.~\ref{equ1}) in our contrastive learning. In this paper, we define 7 ROIs including the whole face, forehead, left top cheek, right top cheek, left bottom cheek, right bottom cheek and chin. 

\subsection{Sparsity-Based Temporal Augmentation}\label{tempAug}

\begin{figure}
\begin{center}
   \includegraphics[width=.8\linewidth]{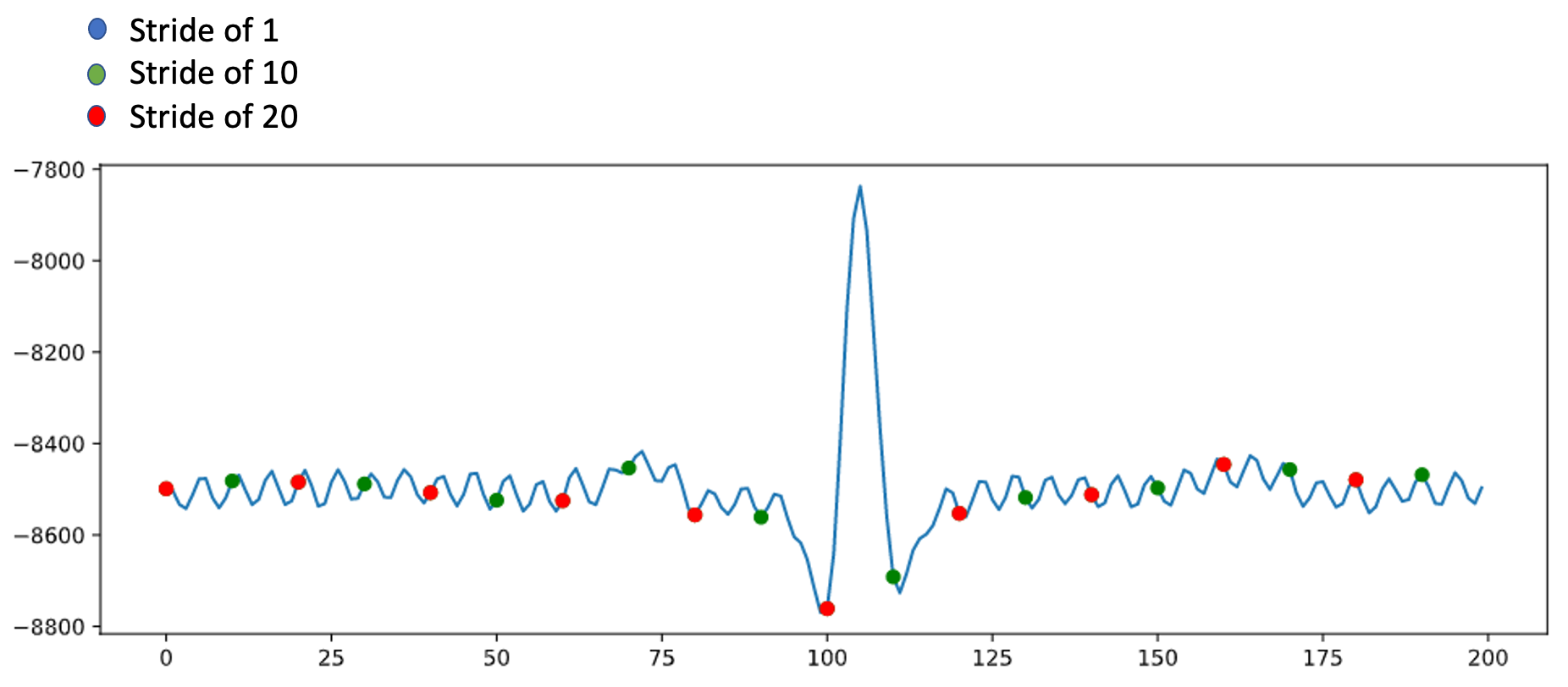}
\end{center}
   \caption{\textbf{Illustration of signal sparsity.} This is an example of 1-second length signal with sampling rate of 256. The \textit{blue} line represents original signal (i.e., stride of $1$), the \textit{green} dots represent data points of stride of $10$, and the \textit{red} dots represent data points of stride of $20$. \textbf{Note:} red points overlap with green points, i.e., green points also appear at red point positions.}
\label{fig:signal}
\end{figure}


The main idea of sparsity-based temporal augmentation is motivated by the physical property of signals as shown in Fig.~\ref{fig:signal}, i.e., Nyquist–Shannon sampling theorem~\cite{NyquistCertainTI} such that a discrete sequence of samples can reconstruct corresponding continuous-time signal if the band-limit $B$ of the signal is less than the sample rate $f_s$, i.e., $B < \frac{1}{2}f_s$. In our case, the Frames Per Second (FPS) can be treated as sampling rate, and as long as the FPS is bigger than the Nyquist rate (i.e., 2 times of rPPG signal band-limit), rPPG signal can be extracted from the video clip. As such, we can use different strides to augment each video along the time-axis, and the larger the stride is the lower the FPS will be.



\subsection{Spatiotemporal Loss with Pseudo-labels}\label{pseudo-label}


Specific constraints such as inductive bias during model training can help extract representative features. It effectively regulates and generalise the domain information \cite{Argyriou2007ConvexMF}. As such, we introduce a spatiotemporal loss that regulates the training process of contrastive learning and handles complicated noise. It first generates pseudo-labels for augmented video clips and creates two additional auxiliary classification tasks to predict the pseudo-labels. The pseudo-labels for each of video clip are defined based on the corresponding data augmentations. Suppose that we have a list of ROIs $\{R_1, ..., R_m, ..., R_M\}$ and a list of strides $\{S_1, ..., S_n, ..., S_N\}$. Then, the video clip augmented by $R_m$ and $S_n$ will be labelled as $(m, n)$ which are used as the ground truths of auxiliary classification tasks. According to previous work \cite{Chen2020ASF}, the mutual information among different views are maximised by 
\begin{equation}
    \mathcal{L}_{[i, j]} = -\log \frac{\exp{(sim(z_i, z_j) / \tau)}}{\sum_{k=0}^{2N} \mathbbm{1}_{k \neq i} \exp{(sim(z_i, z_k) / \tau)}}
    \label{equ1}
\end{equation}
where $(i, j)$ is a pair of positive samples, $\mathbbm{1}_{k \neq i} \in  \{ 0, 1 \}$ is an indicator function which equals to 1 iff $k \neq i$ (i.e., not the feature vector itself) and $\tau$ is the temperature hyper-parameter. Additionally, we generalise this learning by introducing 
\begin{equation}
    \mathcal{L}_a({\boldsymbol y_a}, \hat{\boldsymbol y_a}) = - \sum^{\textrm{C}_{M}}_{i=1} y_a^{i}\log(\hat{y}_a^{i})
    \label{eq6}
\end{equation}
where ${\boldsymbol y_a}$ is the predicted ROIs, $\hat{\boldsymbol y_a}$ is the target ROIs and $\textrm{C}_M$ is the number of classes, and
\begin{equation}
    \mathcal{L}_b({\boldsymbol y_b}, \hat{\boldsymbol y_b}) = - \sum^{\textrm{C}_{N}}_{i=1} y_b^{i}\log(\hat{y}_b^{i})
    \label{eq7}
\end{equation}
where ${\boldsymbol y_b}$ is the predicted strides, $\hat{\boldsymbol y_b}$ is the target strides and $\textrm{C}_N$ is the number of classes. Our spatiotemporal loss is the sum of all losses as follows:
\begin{equation}
    \mathcal{L}_i = \mathcal{L}_{[i, j]} + \mathcal{L}_a + \mathcal{L}_b
    \label{eq8}
\end{equation}
We show the effectiveness of proposed loss function in Sec.\ref{ablation_study}.



\begin{table*}[]
\centering
\resizebox{.8\textwidth}{!}{%
\begin{tabular}{ccclcccc}
\hline
\multicolumn{1}{c|}{\multirow{2}{*}{Strategy}}                                                        & \multicolumn{1}{c|}{\multirow{2}{*}{Method}}            & \multicolumn{2}{c|}{\multirow{2}{*}{Dataset}}    & \multicolumn{4}{c}{HR (bpm)}                                                                                                 \\
\multicolumn{1}{c|}{}                                                                                 & \multicolumn{1}{c|}{}                                   & \multicolumn{2}{c|}{}                            & SD                                 & MAE                                & RMSE                               & R             \\ \hline
\multicolumn{1}{c|}{\multirow{12}{*}{\begin{tabular}[c]{@{}c@{}}Self-\\ Supervised\end{tabular}}}     & \multicolumn{1}{c|}{DPC}                                & \multicolumn{2}{c|}{\multirow{4}{*}{MAHNOB-HCI}} & \multicolumn{1}{c|}{11.76}         & \multicolumn{1}{c|}{9.16}          & \multicolumn{1}{c|}{14.54}         & -0.35         \\
\multicolumn{1}{c|}{}                                                                                 & \multicolumn{1}{c|}{MemDPC}                             & \multicolumn{2}{c|}{}                            & \multicolumn{1}{c|}{10.83}         & \multicolumn{1}{c|}{8.23}          & \multicolumn{1}{c|}{12.14}         & 0.45          \\
\multicolumn{1}{c|}{}                                                                                 & \multicolumn{1}{c|}{SeCo}                               & \multicolumn{2}{c|}{}                            & \multicolumn{1}{c|}{9.48}          & \multicolumn{1}{c|}{7.03}          & \multicolumn{1}{c|}{10.21}         & 0.67          \\
\multicolumn{1}{c|}{}                                                                                 & \multicolumn{1}{c|}{SLF-RPM (ours)}                     & \multicolumn{2}{c|}{}                            & \multicolumn{1}{c|}{{\ul 4.58}}    & \multicolumn{1}{c|}{{\ul 3.60}}    & \multicolumn{1}{c|}{{\ul 4.67}}    & {\ul 0.92}    \\ \cline{2-8} 
\multicolumn{1}{c|}{}                                                                                 & \multicolumn{1}{c|}{DPC}                                & \multicolumn{2}{c|}{\multirow{4}{*}{VIPL-HR-V2}} & \multicolumn{1}{c|}{18.56}         & \multicolumn{1}{c|}{14.07}         & \multicolumn{1}{c|}{19.20}         & -0.45         \\
\multicolumn{1}{c|}{}                                                                                 & \multicolumn{1}{c|}{MemDPC}                             & \multicolumn{2}{c|}{}                            & \multicolumn{1}{c|}{18.03}         & \multicolumn{1}{c|}{13.65}         & \multicolumn{1}{c|}{18.12}         & 0.13          \\
\multicolumn{1}{c|}{}                                                                                 & \multicolumn{1}{c|}{SeCo}                               & \multicolumn{2}{c|}{}                            & \multicolumn{1}{c|}{{\ul 16.56}}   & \multicolumn{1}{c|}{13.32}         & \multicolumn{1}{c|}{{\ul 16.58}}   & 0.23          \\
\multicolumn{1}{c|}{}                                                                                 & \multicolumn{1}{c|}{SLF-RPM (ours)}                     & \multicolumn{2}{c|}{}                            & \multicolumn{1}{c|}{16.60}         & \multicolumn{1}{c|}{{\ul 12.56}}   & \multicolumn{1}{c|}{16.59}         & {\ul 0.32}    \\ \cline{2-8} 
\multicolumn{1}{c|}{}                                                                                 & \multicolumn{1}{c|}{DPC}                                & \multicolumn{2}{c|}{\multirow{4}{*}{UBFC-rPPG}}  & \multicolumn{1}{c|}{11.62}         & \multicolumn{1}{c|}{10.60}         & \multicolumn{1}{c|}{11.92}         & -0.32         \\
\multicolumn{1}{c|}{}                                                                                 & \multicolumn{1}{c|}{MemDPC}                             & \multicolumn{2}{c|}{}                            & \multicolumn{1}{c|}{12.42}         & \multicolumn{1}{c|}{10.85}         & \multicolumn{1}{c|}{12.81}         & 0.25          \\
\multicolumn{1}{c|}{}                                                                                 & \multicolumn{1}{c|}{SeCo}                               & \multicolumn{2}{c|}{}                            & \multicolumn{1}{c|}{9.74}          & \multicolumn{1}{c|}{9.83}          & \multicolumn{1}{c|}{10.62}         & 0.58          \\
\multicolumn{1}{c|}{}                                                                                 & \multicolumn{1}{c|}{SLF-RPM(ours)}                      & \multicolumn{2}{c|}{}                            & \multicolumn{1}{c|}{{\ul 9.60}}    & \multicolumn{1}{c|}{{\ul 8.39}}    & \multicolumn{1}{c|}{{\ul 9.70}}    & {\ul 0.70}    \\ \hline
\multicolumn{8}{c}{}                                                                                                                                                                                                                                                                                                                              \\ \hline
\multicolumn{1}{c|}{\multirow{12}{*}{\begin{tabular}[c]{@{}c@{}}Supervised \\ Learning\end{tabular}}} & \multicolumn{1}{c|}{DeepPhys~\cite{Chen2018DeepPhysVP}} & \multicolumn{2}{c|}{\multirow{5}{*}{MAHNOB-HCI}} & \multicolumn{1}{c|}{-}             & \multicolumn{1}{c|}{4.57}          & \multicolumn{1}{c|}{-}             & -             \\
\multicolumn{1}{c|}{}                                                                                 & \multicolumn{1}{c|}{STVEN + rPPGNet~\cite{Yu3}}         & \multicolumn{2}{c|}{}                            & \multicolumn{1}{c|}{5.57}          & \multicolumn{1}{c|}{4.03}          & \multicolumn{1}{c|}{5.93}          & \textbf{0.88} \\
\multicolumn{1}{c|}{}                                                                                 & \multicolumn{1}{c|}{AutoHR~\cite{Yu2}}                  & \multicolumn{2}{c|}{}                            & \multicolumn{1}{c|}{\textbf{4.73}} & \multicolumn{1}{c|}{3.78}          & \multicolumn{1}{c|}{5.10}          & 0.86          \\
\multicolumn{1}{c|}{}                                                                                 & \multicolumn{1}{c|}{Meta-rPPG (Proto+synth)~\cite{Lee}} & \multicolumn{2}{c|}{}                            & \multicolumn{1}{c|}{4.90}          & \multicolumn{1}{c|}{\textbf{3.01}} & \multicolumn{1}{c|}{\textbf{3.68}} & 0.85          \\
\multicolumn{1}{c|}{}                                                                                 & \multicolumn{1}{c|}{Supervised Baseline (3D ResNet-18)} & \multicolumn{2}{c|}{}                            & \multicolumn{1}{c|}{9.81}          & \multicolumn{1}{c|}{7.34}          & \multicolumn{1}{c|}{9.76}          & 0.56          \\ \cline{2-8} 
\multicolumn{1}{c|}{}                                                                                 & \multicolumn{1}{c|}{RePSS Team 1~\cite{Li2020The1C}}    & \multicolumn{2}{c|}{\multirow{3}{*}{VIPL-HR-V2}} & \multicolumn{1}{c|}{-}             & \multicolumn{1}{c|}{\textbf{8.50}} & \multicolumn{1}{c|}{\textbf{-}}    & -             \\
\multicolumn{1}{c|}{}                                                                                 & \multicolumn{1}{c|}{RePSS Team 5~\cite{Li2020The1C}}    & \multicolumn{2}{c|}{}                            & \multicolumn{1}{c|}{-}             & \multicolumn{1}{c|}{12.00}         & \multicolumn{1}{c|}{-}             & -             \\
\multicolumn{1}{c|}{}                                                                                 & \multicolumn{1}{c|}{Supervised Baseline (3D ResNet-18)} & \multicolumn{2}{c|}{}                            & \multicolumn{1}{c|}{16.69}         & \multicolumn{1}{c|}{12.03}         & \multicolumn{1}{c|}{16.68}         & 0.37          \\ \cline{2-8} 
\multicolumn{1}{c|}{}                                                                                 & \multicolumn{1}{c|}{POS~\cite{wang2016algorithmic}}     & \multicolumn{2}{c|}{\multirow{4}{*}{UBFC-rPPG}}  & \multicolumn{1}{c|}{10.40}         & \multicolumn{1}{c|}{\textbf{4.12}} & \multicolumn{1}{c|}{10.5}          & -             \\
\multicolumn{1}{c|}{}                                                                                 & \multicolumn{1}{c|}{3D CNN~\cite{Bousefsaf20193DCN}}    & \multicolumn{2}{c|}{}                            & \multicolumn{1}{c|}{8.55}          & \multicolumn{1}{c|}{5.45}          & \multicolumn{1}{c|}{8.64}          & -             \\
\multicolumn{1}{c|}{}                                                                                 & \multicolumn{1}{c|}{Meta-rPPG (Proto+synth)~\cite{Lee}} & \multicolumn{2}{c|}{}                            & \multicolumn{1}{c|}{\textbf{7.12}} & \multicolumn{1}{c|}{5.97}          & \multicolumn{1}{c|}{\textbf{7.42}} & \textbf{0.53} \\
\multicolumn{1}{c|}{}                                                                                 & \multicolumn{1}{c|}{Supervised Baseline (3D ResNet-18)} & \multicolumn{2}{c|}{}                            & \multicolumn{1}{c|}{9.68}          & \multicolumn{1}{c|}{8.08}          & \multicolumn{1}{c|}{9.81}          & \textbf{0.53} \\ \hline
\end{tabular}%
}
\caption{\textbf{Linear Evaluation and Supervised Results.} The upper section of Table shows the results of SOTA SSL methods and our SLF-RPM on three datasets. The best performing results for each dataset are underlined. The bottom section of Table shows the results of SOTA supervised HR estimation methods and the supervised baseline (i.e., 3D ResNet-18). The best performing results from each dataset are in bold. }
\label{tab:linear}
\end{table*}
\section{Experiment Setup}\label{experiment}

\textbf{Datasets.} We evaluated our RPM framework on HR estimation task by three widely used public datasets : MAHNOB-HCI~\cite{Soleymani2012AMD}, VIPL-HR-V2~\cite{Li2020The1C} and UBFC-rPPG~\cite{Bobbia2019UnsupervisedST} (See appendix~\ref{dataset}). All videos were re-sampled into 30 FPS and frames were resized into $64 \times 64$, and the length of each clip was constrained into 5 seconds.

\noindent \textbf{Metrics.} The model performance on HR estimation (downstream task of RPM) was measured by comparing metrics of standard deviation (SD), the mean absolute error (MAE), the root mean square error (RMSE) and the Pearson’s correlation coefﬁcient (R). During the evaluation process, we adopt subject-exclusive test~\cite{Niu2}, i.e., subjects in the training set will not appear in the testing set for fair comparisons.

\noindent \textbf{Linear Classification.} To evaluate the quality of extracted rPPG representations, we followed the common linear classification protocol~\cite{Chen2020ASF, He2020MomentumCF} which freezes the weights of self-supervised video encoder layers and trains a subsequent FC layer on the global average pooling features of the video encoder. This evaluation is also used in our ablation studies.

\noindent \textbf{Transfer Learning.} We firstly pre-trained SLF-RPM without labels, and then fine-tuned the weights of video encoder layers and a linear classifier with labels. We compared our method with 3D ResNet-18 pre-trained using Kinetics-700 \cite{Smaira2020ASN} and VIPL-HR-V2 (i.e., largest among three datasets) for supervised pre-training. 


\noindent \textbf{Augmentation Details.} Each video was first augmented into two clips using the stride number randomly sampled from the list $\{1, 2, 3, 4, 5\}$\footnote{Given the range of human HR is 40-160 beat per minute, the biggest stride for 30-fps videos we can use is 5 which generates 6-fps clips (the minimum sampling rate for recovering HR signals).}, and each augmented clip was constrained to have length of 30-frame. Therefore, the longest clip (i.e., stride of 5) contained 5-second information, while the shortest clip (i.e., stride of 1) had 1-second information. Frames of each clip were then cropped based on a specific ROI by randomly selecting one from 7 pre-defined facial areas as described in Sec.~\ref{spatialAug}. 

We used Adam optimiser \cite{Kingma2015AdamAM} with default settings to update the model weights. Settings of hyper-parameters can be found in appendix~\ref{params}.

\section{Results and Discussions}

\subsection{Linear Classification Evaluation}\label{Sec:linear}

The results of linear classification experiment are shown in Table~\ref{tab:linear}. We re-implemented three SOTA video SSL methods including DPC~\cite{Han2019VideoRL}, MemDPC~\cite{Han2020MemoryaugmentedDP} and SeCo~\cite{Yao2020SeCoES}. Our method had the best accuracies relative to other SSL methods on all datasets. SeCo was the closest to our method achieving MAE of 7.03 on MAHNOB-HCI, 13.32 on VIPL-HR-V2 and 9.83 on UBFC-rPPG. We attribute this difference to our augmentation schemes that effectively captured subtle colour fluctuations on facial skin, whereas other SSL methods are limited to learn only from apparent motions of the objects.

Performance comparisons to the supervised methods are also shown in Table~\ref{tab:linear}. Our method on MAHNOB-HCI had a better performance (MAE of 3.60) with a large margin than the Supervised Baseline (MAE of 7.34) and had superior accuracy than previous SOTA Deephys \cite{Chen2018DeepPhysVP} (MAE of 4.57) and rPPGNet \cite{Yu3} (MAE of 4.03). Comparing with the best performing supervised approach Meta-rPPG \cite{Lee} (MAE of 3.01) which used unlabelled testing samples to conduct transductive meta-learning, our method showed a competitive result and achieved higher SD of 4.58 and R of 0.92, suggesting that our method had more consistent predictions and stronger linear correlation between ground truth and predictions. We suggest that our distinctive performance on MAHNOB-HCI over supervised methods is subject to the limitations of the dataset where: 1. number of participants in MAHNOB-HCI is limited (i.e., 27 subjects in total); 2. videos are highly compressed, losing subtle details that human eyes cannot see. This was also reported from other similar work \cite{Yu3}. Nevertheless, our method implicitly modelled compression-corrupted information patterns and learned robust rPPG signals in a self-supervised manner.

Our SLF-RPM on UBFC-rPPG achieved MAE of 8.39. Supervised methods on UBFC-rPPG had better performances than other SSL methods.This is mainly attributed to less noise on UBFC-rPPG (uncompressed video data), which improved overall supervised learning outcomes. Nevertheless, SLF-RPM reduced the performance gap with Supervised Baseline (MAE of 8.08) to MAE of 0.31. 

The performances of all the methods on VIPL-HR-V2 were reduced in comparison to other two datasets. This is because VIPL-HR-V2 has more complex video conditions. Nonetheless, our method (MAE of 12.56) successfully closed the performance gap with the best supervised approach RePSS Team 1 \cite{Li2020The1C} (MAE of 8.5) and had competitive accuracy with Supervised Baseline (MAE of 12.03).


\subsection{Transfer Learning}\label{transfer}

We evaluated the transferable ability of rPPG features extracted by SLF-RPM. The results of the transfer learning are shown in Table~\ref{tab:transfer}. Our results indicate that the SLF-RPM improved the estimation of HR when it was used to pre-train CNNs for subsequent supervised fine-tuning. When the available dataset is limited, SLF-RPM can be used as the self-pre-training strategy (i.e., pre-train and fine-tune on the same dataset). This enabled faster adaption for rPPG feature extraction such that MAE was reduced to 11.59 on VIPL-HR-V2 dataset. Moreover, it was most effective when SLF-RPM was applied to pre-train using the largest VIPL-HR-V2 dataset. The performance of Supervised Baseline was improved from MAE of 7.34 to 6.23 on MAHNOB-HCI and from MAE of 8.08 to 7.35 on UBFC-rPPG dataset.We suggest that our self-supervised pre-training acted as an initialisation point for effective supervised fine-tuning, which improved the feature representation of rPPG signals.

In contrast, model pre-trained on Kinetics-700 benchmark, commonly used in action recognition task, degraded HR estimation performance (MAE of 8.98 on MAHNOB-HCI, 12.28 on VIPL-HR-V2 and 9.77 on UBFC-rPPG) compared with the baseline model due to the hard domain shift between tasks. Moreover, we evaluated the model pre-trained with labels using VIPL-HR-V2 and achieved MAE of 8.06 on MAHNOB-HCI and 8.32 on UBFC-rPPG. This demonstrates that our self-supervised pre-training strategy was better at characterising the RPM features than that of supervised pre-training methods. 





\begin{table*}[]
\centering
\resizebox{.8\textwidth}{!}{%
\begin{tabular}{c|c|c|c|cccc}
\hline
\multirow{2}{*}{Strategy}                                                     & \multirow{2}{*}{Method}                  & \multirow{2}{*}{Pre-training Dataset} & \multirow{2}{*}{Fine-tuning Dataset} & \multicolumn{4}{c}{HR (bpm)}                                                                                                    \\
                                                                              &                                          &                                       &                                      & SD                                  & MAE                                 & RMSE                                & R             \\ \hline
\multirow{8}{*}{\begin{tabular}[c]{@{}c@{}}Transfer\\  Learning\end{tabular}} & \multirow{5}{*}{Supervised-Pre-Training} & \multirow{3}{*}{Kinetics-700}         & MAHNOB-HCI                           & \multicolumn{1}{c|}{11.89}          & \multicolumn{1}{c|}{8.98}           & \multicolumn{1}{c|}{13.03}          & -0.07         \\
                                                                              &                                          &                                       & VIPL-HR-V2                           & \multicolumn{1}{c|}{16.82}          & \multicolumn{1}{c|}{12.28}          & \multicolumn{1}{c|}{16.86}          & 0.35          \\
                                                                              &                                          &                                       & UBFC-rPPG                            & \multicolumn{1}{c|}{11.44}          & \multicolumn{1}{c|}{9.77}           & \multicolumn{1}{c|}{11.79}          & 0.52          \\ \cline{3-8} 
                                                                              &                                          & \multirow{2}{*}{VIPL-HR-V2}           & MAHNOB-HCI                           & \multicolumn{1}{c|}{10.89}          & \multicolumn{1}{c|}{8.06}           & \multicolumn{1}{c|}{11.90}          & \textbf{0.72} \\
                                                                              &                                          &                                       & UBFC-rPPG                            & \multicolumn{1}{c|}{10.87}          & \multicolumn{1}{c|}{8.32}           & \multicolumn{1}{c|}{10.87}          & 0.61          \\ \cline{2-8} 
                                                                              & \multirow{3}{*}{SLF-RPM}                 & \multirow{3}{*}{VIPL-HR-V2}           & MAHNOB-HCI                           & \multicolumn{1}{c|}{\textbf{10.19}} & \multicolumn{1}{c|}{\textbf{6.23}}  & \multicolumn{1}{c|}{\textbf{10.35}} & 0.56          \\
                                                                              &                                          &                                       & VIPL-HR-V2                           & \multicolumn{1}{c|}{\textbf{15.55}} & \multicolumn{1}{c|}{\textbf{11.59}} & \multicolumn{1}{c|}{\textbf{15.60}} & \textbf{0.46} \\
                                                                              &                                          &                                       & UBFC-rPPG                            & \multicolumn{1}{c|}{\textbf{10.19}} & \multicolumn{1}{c|}{\textbf{7.35}}  & \multicolumn{1}{c|}{\textbf{10.53}} & \textbf{0.63} \\ \hline
\end{tabular}%
}
\caption{\textbf{Transfer Learning Results.} This table shows model performances under different pre-training strategies. The transfer abilities of learned representations are evaluated on three datasets. The best performing results from each dataset are in bold.}
\label{tab:transfer}
\end{table*}

\subsection{Ablation Studies}\label{ablation_study}

\noindent \textbf{Landmark-based Spatial Augmentation}

We compared our landmark-based spatial augmentation with 5 other common data augmentation techniques. We applied these techniques to the whole face area (i.e., $R_1$ of the pre-defined ROI list) and used stride of 1 to avoid temporal augmentation effects. The descriptions of $R_1-R_7$ are in Sec.~\ref{spatialAug}. As shown in Table~\ref{tab:ablationSpatial}, our landmark-based spatial augmentation had the best MAE score of 5.12. Our results show that the \textit{whole face} ROI ($R_1$) was effective in model training, which improved MAE from 7.98 to 5.12. Among the standard augmentation techniques, \textit{Random Crop and Resize} had the best result with MAE of 6.74. \textit{Random Grayscale} had the worst performance with MAE of 10.66. The combination of all 5 techniques (standard paradigm in simCLR\cite{Chen2020ASF}), in fact, reduced the performance giving MAE of 9.54. One possible reason is that although appearance transformations prevent the model from using colour histogram shortcut~\cite{Chen2020ASF} to distinguish different views, signal-related information contained in colour channels can be also distorted. Another reason could be that geometric transformations cannot guarantee augmented inputs are valuable. They may introduce signal noise by including non-facial areas \cite{Li2014RemoteHR}. Overall, our landmark-based spatial augmentation outperformed standard spatial augmentation techniques by a large margin for the task of HR estimation.

\begin{table}[]
\centering
\resizebox{0.7\linewidth}{!}{%
\begin{tabular}{c|c}
\hline
\multirow{2}{*}{Method}                      & HR (bpm)           \\
                                             & MAE                \\ \hline
Random Crop and Resize                       & 6.74               \\
Random Horizontal Flip                       & 6.99               \\
Colour Jitter                                & 8.11               \\
Random Grayscale                             & 10.66              \\
Gaussian Blur                                & 9.81               \\
Combined Above 5 Augmentations               & 9.54               \\ \hline
\{$R_2, R_3, R_4, R_5, R_6, R_7$\}                     & 7.98             \\
\{$R_1, R_3, R_4, R_5, R_6, R_7$\}                     & 6.51             \\
\{$R_1, R_2, R_4, R_5, R_6, R_7$\}                     & 7.03             \\
\{$R_1, R_2, R_3, R_5, R_6, R_7$\}                     & 7.15             \\
\{$R_1, R_2, R_3, R_4, R_6, R_7$\}                     & 6.45             \\
\{$R_1, R_2, R_3, R_4, R_5, R_7$\}                     & 5.82             \\
\{$R_1, R_2, R_3, R_4, R_5, R_6$\}                     & 6.72             \\ \hline
\{$R_1, R_2, R_3, R_4, R_5, R_6, R_7$\}                 & \textbf{5.12}    \\ \hline
\end{tabular}%
}
\caption{\textbf{Ablation on spatial augmentation.} We compared different spatial augmentations and evaluated the impact of each ROI. The best result is in bold. No temporal augmentation is applied.}
\label{tab:ablationSpatial}
\end{table}

\noindent \textbf{Sparsity-based Temporal Augmentation}

We evaluated the effectiveness of our sparsity-based temporal augmentation compared to different standard temporal augmentation techniques, and use \textit{whole face} $R_0$ to avoid spatial augmentation effects. As shown in the bottom part of Table~\ref{tab:abalationTemporal}, our sparsity-based temporal augmentation follows the Nyquist-Shannon sampling theorem and showed the larger sparsity ranges had under-sampling issue that negatively affect the learning outcome since sparsity range \{1,2,3,4,5,6\} had worse MAE of 5.92 than the best performing range \{1,2,3,4,5\} with MAE of 5.28. However, we also noted that we need some enough sparsity (e.g., \{1,2,3,4,5\}) to have stronger data transformation compared with smaller range \{1,2,3,4\} achieving MAE of 6.21.

Nevertheless, the proposed sparsity-based temporal augmentation generally outperformed other augmentation techniques. Among these four standard temporal augmentations, \textit{Random Temporal Interval} had the best result with MAE of 5.74 and we attribute this to its ability to characterise the periodic cycle of blood volume changes on facial skin. \textit{Periodic} had the worst MAE score of 7.31 due to the distorted rPPG signals.


\begin{table}[]
\centering
\resizebox{0.9\linewidth}{!}{%
\begin{tabular}{c|c}
\hline
\multirow{2}{*}{Method}                                         & HR (bpm)          \\
                                                                & MAE               \\ \hline
Random Temporal Interval~\cite{Qian2020SpatiotemporalCV}        & 5.74              \\
Random Permutation~\cite{Jenni2020VideoRL}                      & 7.18              \\
Periodic~\cite{Jenni2020VideoRL}                               & 7.31              \\
Warp~\cite{Jenni2020VideoRL}                                   & 7.22              \\ \hline
Sparsity range from \{1, 2, 3, 4\}                             & 6.21     \\
Sparsity range from \{1, 2, 3, 4, 5\}                          & \textbf{5.28}     \\ 
Sparsity range from \{1, 2, 3, 4, 5, 6\}                       & 5.92     \\ \hline
\end{tabular}%
}
\caption{\textbf{Ablation on temporal augmentation.} We compared with 4 standard temporal augmentation techniques proposed in previous studies. The best result is in bold. No spatial augmentation is applied.}
\label{tab:abalationTemporal}
\end{table}

\noindent \textbf{Effect of Spatiotemporal Loss with Pseudo-labels }

\begin{table}[]
\centering
\resizebox{0.7\linewidth}{!}{%
\begin{tabular}{c|c}
\hline
\multirow{2}{*}{Method}                & HR (bpm) \\
                                       & MAE      \\ \hline
SLF-RPM without pseudo-labels & 4.25     \\
SLF-RPM with pseudo-labels    & \textbf{3.6}      \\ \hline
\end{tabular}%
}
\caption{\textbf{Ablation on spatiotemporal loss with pseudo-labels.} We compared model performance with and without pseudo-labels integration. The best result is in bold.}
\label{tab:classIntegra}
\end{table}

To validate the effectiveness of our spatiotemporal loss, we compared the model performance under two different scenarios, i.e., with and without pseudo-labels integration. From Table~\ref{tab:classIntegra}, our model, by additionally assigning two classification tasks, improved the overall performance achieving the MAE of 3.6. We suggest that this is because pseudo-labels enabled better characterisation of complicated ROIs and subtle temporal changes (i.e., noise).



\noindent \textbf{Different Contrastive Learning Strategy}

As the first contrastive learning work for RPM, we investigated the performance of different SSL strategies including CPC \cite{Oord2018RepresentationLW}, MoCo \cite{He2020MomentumCF}, CPC + MoCo and simCLR \cite{Chen2020ASF} which achieved SOTA results on image classification tasks. We extended their ideas into video applications, i.e., transforming from 2D data into 3D data. To make our experiments consistent, data augmentations were limited to spatial augmentations using the same combination adopted in simCLR. From Table~\ref{tab:contrastStrategy}, simCLR obtained best result (MAE of 9.54), whereas CPC only achieved MAE of 13.49.

\begin{table}[]
\centering
\resizebox{0.6\linewidth}{!}{%
\begin{tabular}{c|c}
\hline
\multirow{2}{*}{Contrastive Learning Strategy} & \multirow{2}{*}{\begin{tabular}[c]{@{}c@{}}HR(bpm)\\ MAE\end{tabular}} \\
                                               &                                                                        \\ \hline
CPC \cite{Oord2018RepresentationLW}            & 13.49                                                                  \\
CPC + MoCo              & 13.05                                                                  \\
MoCo    \cite{He2020MomentumCF}                              & 11.73                                                                  \\
simCLR \cite{Chen2020ASF}                      & \textbf{9.54}                                                                   \\ \hline
\end{tabular}%
}
\caption{\textbf{Ablation on contrastive learning strategy.} We compared with SOTA contrastive learning strategies for the HR estimation task. The best result is in bold.}
\label{tab:contrastStrategy}
\end{table}

\section{Conclusion}

We present a SSL framework for RPM by introducing novel landmark-based spatial augmentation, sparsity-based temporal augmentation and spatiotemporal loss. Our results showed that the SLF-RPM significantly outperformed other SSL methods and achieved a competitive accuracy compared to other supervised methods on HR estimation task. The superior transfer ability of learnt RPM representations using SLF-RPM demonstrates that it can be used as an effective pre-training strategy for many facial video analysis tasks. The limitations of this work is shown in appendix~\ref{limits}



\appendix

\section{SLF-RPM Algorithm}\label{alg}

We demonstrates the overall algorithm for SLF-RPM in Alg.~\ref{alg1}.

\begin{algorithm}{}

\newcommand\mycommfont[1]{\small\ttfamily{#1}}
\SetCommentSty{mycommfont}

\newlength\mylen
\newcommand\myinput[1]{%
  \settowidth\mylen{\KwIn{}}%
  \setlength\hangindent{\mylen}%
  \hspace*{\mylen}#1\\}

\SetAlgoLined 
    \KwIn{Videos $V = \{v_1, v_2, ..., v_N\}$ with $N$ videos and each video frame consists of skin pixels $C_k(t)$ (Sec.3.3); \newline
    Stride list $S = \{s_1, s_2, ..., s_M\}$ with $M$ values;\newline
    ROI list $R = \{r_1, r_2, ..., r_L\}$ with $L$ values; Clip frame number $D$.}
    
    \DontPrintSemicolon
    \SetKwFunction{FMain}{$f_1$}
    \SetKwProg{Fn}{Spatial Augmentation}{:}{}
    \Fn{\FMain{$v_i$}}{
          Randomly select a value $r_i$ from $R$, and generate a clip using ROI $r_i$ only from $v_i$.\;
    }
    \;
  
    \SetKwFunction{PMain}{$f_2$}
    \SetKwProg{Pn}{Temporal Augmentation}{:}{}
    \Pn{\PMain{$v_i$}}{
          Randomly select a value $s_i$ from $S$, and generate a $D$-frame clip with stride of $s_i$ from $v_i$.\;
    }
    \;
    
    \SetKwFunction{PMain}{$f_3$}
    \SetKwProg{Pn}{Contrastive Loss}{:}{}
    \Pn{\PMain{$V$}}{
          Given a video set $V$ containing positive and negative samples, conduct contrastive learning and return the loss.\;
    }
    \;
    
    \SetKwFunction{PMain}{$f_4$}
    \SetKwProg{Pn}{Spatiotemporal Loss}{:}{}
    \Pn{\PMain{$V,\ A $}}{
          Given a video set $V$ and corresponding augmentation list $A$, classify each video's transformation and return the loss.\;
    }
    \;
    
    \tcp{Collection of augmented video clips $V'$ and corresponding transformations $A$.}
    
    $V' = \{\}$
    
    $A = \{\}$
    
    \tcp{Spatiotemporal augmentation to generate different views.}
    
    \For{$k \gets {1, ..., N}$}{
        \tcp{1st spatiotemporal augmentation to get augmented $c_k^1$.}
        $c_k^1, s_m^1 = f_2(v_k)$
        
        $c_k^1, r_l^1 = f_1(c_k^1)$
        
        \tcp{2nd spatiotemporal augmentation to get augmented $c_k^2$.}
        
        $c_k^2, s_m^2 = f_2(v_k)$
        
        $c_k^2, r_l^2 = f_1(c_k^2)$
        
        $A\ \text{append}\ \{(s_m^1, r_l^1), (s_m^2, r_l^2)\}$
        
        $V'\ \text{append}\ \{(c_k^1, c_k^2)\}$
    }
    
    \tcp{$A = \{(s_m^1, r_l^1), (s_m^2, r_l^2), ..., (s_m^1, r_l^1), (s_m^2, r_l^2)\}$, $m,l$ are random numbers.}
    
    \tcp{$V' = \{c_1^1, c_1^2, ..., c_{N}^1, c_{N}^2\}$.}
    
    $loss = f_3(V') + f_4(V', A)$
    
    \KwOut{Sum of contrastive loss and spatiotemporal loss $loss$.}
    
    \caption{SLF-RPM Algorithm}
    \label{alg1}
\end{algorithm}

\section{Dataset Descriptions}\label{dataset}

\textbf{MAHNOB-HCI Dataset} contains 27 subjects, including 12 males and 15 females. Each video was recorded in $780 \times 580$ resolution with frame per second (FPS) of 61. Following the previous settings \cite{Li2014RemoteHR}, we used RGB videos recorded in \textit{emotion elicitation experiment} and selected continuous sequence of frames in the range from 306 to 2135 (around 30 seconds), which resulted in 527 videos. The ground truth HRs were calculated from ECG signals in the channel \textit{EXG2}.

\textbf{UBFC-rPPG Dataset} contains 42 uncompressed videos of 30 FPS from 42 subjects. Ground truth HRs are provided for each of videos. Since this dataset has limited number of videos, we augmented each video by using a sliding window of 150 frames without any frame overlap, i.e., 60 seconds video was split into 12 clips with the length of 5 seconds each. We obtained 523 videos after the augmentation.

\textbf{VIPL-HR-V2 Dataset} is a recently released large-scale HR estimation benchmark which contains 2,500 RGB videos of 500 subjects recorded in $960 \times 720$ resolution with FPS of 30 in average. This dataset is more challenging as it introduces more complicated scenarios including various head movements, different illumination conditions and wide range of HR values \cite{Niu3}. Ground truth HRs and FPS are provided for each video. 




\section{Model Parameters}\label{params}

\textbf{Self-Supervised Training.} For MAHNOB-HCI dataset, the hyper-parameter $\tau$ was 1, output dimension of MLP with 2048, learning rate of 1e-4, batch size of 128, and epoch number with 150. For UBFC-rPPG, the hyper-parameter $\tau$ was set to 0.1, output dimension of MLP with 2048, learning rate of 1e-4, batch size of 128, and epoch number with 50. For VIPL-HR-V2 dataset, the hyper-parameter $\tau$ was 1, output dimension of MLP with 512, learning rate of 1e-5, batch size of 128, and epoch number with 200. We empirically derived the best performing $\tau$ for each dataset.

\textbf{Linear Classification Evaluation.} We aligned the training process with self-supervised settings by sampling 75-frame clips with the stride of 2 from 150-frame videos. The learning rate was set to 5e-3, and the epoch number was set to 50 for MAHNOB-HCI as well as UBFC-rPPG and 100 for VIPL-HR-V2.

\textbf{Transfer Learning.} For MAHNOB-HCI dataset, the epoch number was 100, the batch size was 64 and the learning rate was 1e-4. For UBFC-rPPG, the epoch number was 200, the batch size was 64 and the learning rate was 5e-3. For VIPL-HR-V2, the epoch number was 100, the batch size was 64 and the learning rate was 5e-5. Similarly, videos were down-sampled with stride of 2 such that 75-frame clips were fed into the model.

\section{Limitations}\label{limits}

There are several limitations in this work. First of all, the selection of ROIs in our spatial augmentation is based on prior knowledge, which may implicitly ignore underlying patterns of human skin we have not noticed. For example, people have different sensitiveness to the sunlight such that areas in the face exhibit different signal quality for each individual \cite{Rouast2016RemoteHR,Chong2019SkinDI} and extra noise might be introduced if uniform ROI shapes are applied. Based on that, it is more reasonable to build an algorithm to automate this process and thus the selection of ROIs becomes intelligent and specific to each of the face. Moreover, as suggested in AutoAugmentation~\cite{Cubuk2018AutoAugmentLA}, instead of randomly sampling form pre-defined augmentation list, we can build a search space with sub-policies which could decide the most suitable augmentations for current video clip such that rPPG signal information are maximised. Furthermore, to control the effect of data augmentations, extra hyper-parameters can be added to re-weight the importance of applied transformations such that rPPG signal features can be learnt smoothly.

Another limitation is that we did not study the effect of skin tone on RPM and this can be an issue when transferring among datasets which contain different racial majority. To solve this problem, a larger dataset with balanced ratio of ethics should be built.

According to previous self-supervised method \cite{Chen2020ASF}, the batch size is normally large (e.g., 4096) to provide more negative samples. However, due to the limitation of our dataset size, we can only use batch size of 128. We believe the performance of SLF-RPM can be further enhanced if more negative samples are available. Similarly, because of the limited resource of GPU, we did not experiment the effect of frame size and clip length. In this work, we sacrificed bigger frame and longer clip to make batch size as large as possible which can be fed into a single GPU.

Although our method is effective on improving baseline supervised model, the performance on improving SOTA supervised methods were not experimented. This actually can be done by applying SLF-RPM to initialise their backbone parameters.

Lastly, since our experiments were conducted in the laboratory environment, the effectiveness of our method in the clinic usage is unknown. Our next plan is cooperating with the hospital and evaluating our model in a more realistic situation.

\bibliography{aaai22}

\end{document}